\documentclass[10pt,twocolumn,letterpaper]{article}

\usepackage[applications]{wacv}      % To produce the REVIEW version for the applications track
\usepackage{graphicx}
\usepackage{amsmath}
\usepackage{amssymb}
\usepackage{booktabs}
\usepackage{comment}
\usepackage{xcolor}
\usepackage{pifont}
\newcommand*{\affaddr}[1]{#1} % No op here. Customize it for different styles.
\newcommand*{\affmark}[1][*]{\textsuperscript{#1}}

\usepackage[pagebackref,breaklinks,colorlinks]{hyperref}
\usepackage[capitalize]{cleveref}
\crefname{section}{Sec.}{Secs.}
\Crefname{section}{Section}{Sections}
\Crefname{table}{Table}{Tables}
\crefname{table}{Tab.}{Tabs.}

\def\wacvPaperID{239} % *** Enter the WACV Paper ID here
\def\confName{WACV}
\def\confYear{2024}

\begin{document}

%When citing a multi-author paper, you may save space by using ``et alia'', shortened to ``\etal'' (not ``{\em et.\ al.}'' as ``{\em et}'' is a complete word).

% Update the wacv.cls to do the following automatically.
% For this citation style, keep multiple citations in numerical (not
% chronological) order, so prefer \cite{Alpher03,Alpher02,Authors14} to
% \cite{Alpher02,Alpher03,Authors14}.

\title{Driving through the Concept Gridlock: Unraveling Explainability Bottlenecks 
%for Personalized Interventions 
in Automated Driving}

\author{%
Jessica Echterhoff\affmark[1], An Yan\affmark[1], Kyungtae Han\affmark[2], Amr Abdelraouf\affmark[2], Rohit Gupta\affmark[2], Julian McAuley\affmark[1]\\
\affaddr{\affmark[1]University of California, San Diego}\\
%, \{jechterh, ayan, jmcauley\}@ucsd.edu}\\
\affaddr{\affmark[2]InfoTech Labs, Toyota Motor North America R\&D}\\
}

\maketitle

%%%%%%%%% ABSTRACT
\begin{abstract}
Concept bottleneck models have been successfully used for explainable machine learning by encoding information within the model with a set of human-defined concepts. In the context of human-assisted or autonomous driving, explainability models can help user acceptance and understanding of decisions made by the autonomous vehicle, which can be used to rationalize and explain driver or vehicle behavior. We propose a new approach using concept bottlenecks as visual features for control command predictions and explanations of user and vehicle behavior. We learn a human-understandable concept layer that we use to explain sequential driving scenes while learning vehicle control commands. 
%Our method uses an end-to-end prediction method from sequences of images to control commands, and a concept bottleneck within the model to explain driving behavior. 
This approach can then be used to determine whether a change in a preferred gap or steering commands from a human (or autonomous vehicle) is led by an external stimulus or change in preferences. We achieve competitive performance to latent visual features while gaining interpretability within our model setup. \footnote{The code for this work is available at \url{https://github.com/jessicamecht/concept_gridlock}.}
\end{abstract}

%%%%%%%%% BODY TEXTs

\section{Introduction}
Understanding how human drivers and autonomous vehicles make decisions is essential to ensure safe and reliable operation in various real-world scenarios. Neural networks are powerful tools used for automated learning in the field of self-driving cars \cite{sachdeva22gapformer, Li_2023_WACV, Malla_2023_WACV, Noguchi_2023_WACV, spielberg2019neural,garimella2017neural, Bahari_2022_CVPR}. However, one significant challenge associated with deep neural networks is their nature as black-box models, which hinders the interpretability of their decision-making process.
\begin{figure}
    \centering
\includegraphics[width=0.45\textwidth]{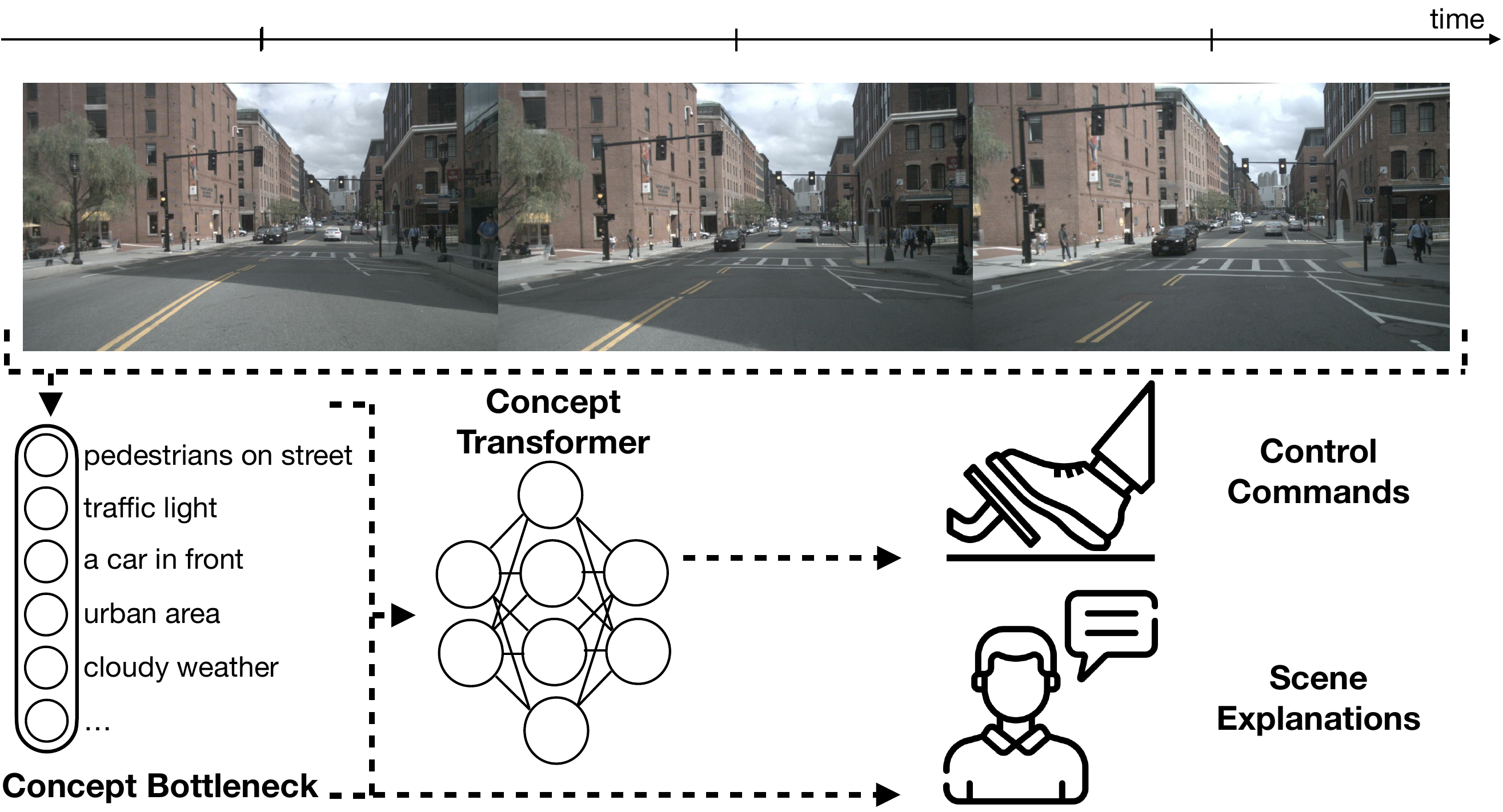}
    \caption{Our proposed framework combines the power of concept bottlenecks and Longformer \cite{beltagy2020longformer} architecture to enable interpretable prediction of control commands in automated driving. By incorporating human-defined concepts within the concept bottleneck layers, we unravel the explainability bottlenecks for safer and more reliable driving. The Longformer architecture allows capturing long-range sequential dependencies in driving scenarios and reveals interesting subsequences through its attention mechanism, while the concept bottlenecks enhance transparency explaining these through driving-related concepts. }
    \label{fig:enter-label}
\end{figure}
This paper proposes to address this challenge by applying concept bottleneck models for explaining driving scenarios. Concept bottleneck models incorporate vision-based human-defined concepts within a bottleneck in the model architecture \cite{oikarinen2023label, koh2020concept}. By encoding driving and scenario-related concepts into the decision-making process, our objective is to provide interpretable and explainable insights into the factors that influence the actions of both drivers and autonomous vehicles. Previous research has demonstrated the effectiveness of learning vehicle controls for autonomous driving \cite{urmson2008autonomous, xu2017end, buehler2009darpa, levinson2011towards, xu2017end}, but the lack interpretability poses challenges to trust, safety, and regulatory compliance. The development of interpretable and explainable models has thus gained significant attention in the research community, aiming to bridge the gap between the performance and interpretability of deep learning models.

Our proposed procedure offers a novel approach to address this interpretability gap in a sequential setup. By incorporating human-defined concepts into the bottleneck of the model architecture, we provide a means to understand and interpret the decision-making process of drivers and autonomous vehicles. Our results can be used for driver intervention prediction in applications such as adaptive cruise control or lane keeping. 

This work provides the following contributions: 
\begin{itemize}
    \item 
    %We use concepts within the model architecture for sequential settings in autonomous driving to gain model and action interpretability and explain driving scenarios via natural language. 
    We propose a novel pipeline for explainable driving that builds concepts with large language models, converts image features into explicit concept scores, and then learns sequential patterns with a Longformer architecture. We provide extensive experiments around model architectures and feature backbones including traditional approaches such as Residual Neural Networks (ResNet), Contrastive Language-Image Pretraining (CLIP) models, and Vision Transformers (ViT) for both single and multi-task setups.
    \item  We find that the concept space maps accurately to different driving conditions and that we can use our transformer attention mechanism to select when to reveal automated system explanations to a driver, highlighting the utility of concept bottleneck models for the rather unexplored sequential settings.
\end{itemize}
Our experimental results demonstrate the effectiveness of concept bottleneck models in sequential learning. Our interpretability analysis reveals that our concept bottleneck models offer insights into the factors influencing a driver's and subsequently a model's decision-making process, enhancing transparency and trustworthiness in autonomous driving systems. For example, we show that they can explain changes in driving behavior such as change of forward distance and give reasoning for those changes.
\section{Related Work}
\subsection{End-to-End Learning of Vehicle Controls}
Research in automated driving has examined perception-based tasks such as finding lane markings,  traffic lights, recognizing traffic participants \cite{urmson2008autonomous,levinson2011towards, buehler2009darpa} as well as end-to-end processes to learn vehicle controls \cite{bojarski2016end, xu2017end}.  Xu et al. \cite{xu2017end} explore a stateful model using a dilated deep neural network and recurrent neural network to predict future vehicle motion given input images. Bojarski et al. \cite{bojarski2016end} train a deep neural network to map front-facing video frames to steering controls.  Hecker et al. \cite{hecker2018end} explore an extension of a model taking multiple modalities
%a surround-view multi-camera system, a route planner, and a CAN bus reader 
as input for control prediction. Different approaches use behavioral cloning to learn a driving policy as a supervised learning problem over observation-action pairs from human driving demonstrations \cite{levinson2011towards}, but only a few explain the rationale for system decisions \cite{kim2018textual}, which makes their behavior opaque and un-interpretable. 
\subsection{Concept Bottleneck Models}
Using human concepts to interpret model behavior has been drawing increasing interest \cite{Kim2017InterpretabilityBF, Bau2017NetworkDQ}. Concept bottleneck models \cite{koh2020concept} extend the idea of first predicting image concepts, then using these concepts to predict a classification target \cite{Lampert2009LearningTD}. Original concept bottleneck models learn the concept space jointly or sequentially with a classification or regression task \cite{koh2020concept}. These models introduce interpretability benefits, but require training the model using concept and class labels, which can be a key limitation. Label-free concept bottleneck models \cite{oikarinen2023label} or models with unsupervised concepts \cite{Sawada2022ConceptBM} alleviate this problem. Most of the work on concept bottleneck models evaluates supervised classification task setups\cite{koh2020concept,oikarinen2023label, Sawada2022ConceptBM,yan2023learning,yan2023robust}. 

The evaluation of concept bottleneck models in sequential settings remains relatively unexplored. But notably, concept bottleneck models enable the identification of key factors or features that can contribute to driving decisions. By extracting concept representations from input data, relevant concepts that are driving the predictions are highlighted. Sequential evaluations provide valuable insights while capturing temporal dependencies and understanding how concepts evolve over time in dynamic scenarios such as driving. Our work focuses on evaluating concept bottleneck models in sequential tasks to assess their performance and interpretability in dynamic decision-making domains.
\subsection{Vehicle Action Explanations}
The importance of explanations for an end-user has been studied from the psychological perspective \cite{lombrozo2006structure, lombrozo2012explanation} indicating the benefit of explanations in autonomous driving. Different work focuses on visual explanations \cite{hendricks2018grounding, hendricks2016generating, kim2020advisable}; \eg Wang et al. \cite{wang2019deep} introduce an instance-level attention model that finds objects that the network needs to pay attention to. Such visual attention may be less convenient (in the driving domain) for users to ``replay". It is therefore important to be able to justify the decisions made and explain why they are reasonable in a convenient manner, \eg in natural language \cite{hendricks2018grounding, hendricks2016generating, kim2020advisable}. Previous research in the field of explainable decision-making in autonomous vehicles explores the use of recurrent neural networks for explanation generation. Kim et al. \cite{kim2018textual} use an architecture based on a convolutional image feature encoder and learn vehicle sensor measurements such as speed while aligning temporal and spatial attention. Their explanation generation process uses an LSTM \cite{10.1162/neco.1997.9.8.1735} to predict next-word probabilities. In contrast, our work demonstrates the potential of concept bottleneck models in providing insights into the decision-making process. To incorporate information on the scene, Kim et al. \cite{kim2019grounding} use an active approach to feed human-to-vehicle advice into the vehicle controller. However, this requires \emph{a priori} information from the human on a situation that is often difficult to obtain. Similarly, Kim et al. \cite{kim2020advisable} propose a system to learn vehicle control with the help of human advice. Those works show that human advice is useful, but do not directly explain why a particular model makes a particular decision.

Despite these advancements, there is still a need for further research to develop robust and effective approaches for explaining driver and autonomous vehicle decisions. Existing studies focus on specific aspects of \emph{post-hoc} explanations or how to use explanations \emph{a priori}, but a framework that integrates human-defined concepts for automated driving \emph{in-situ} within the model to enable white-box model explanations is lacking. Our paper addresses this gap by proposing a novel approach that utilizes concept bottleneck models to encode various driving-related concepts within the decision-making process. By incorporating concepts we aim to provide a holistic understanding of the factors influencing driver and autonomous vehicle actions from within the model.
\begin{figure}[b]
     \centering
         \includegraphics[width=0.4\textwidth]{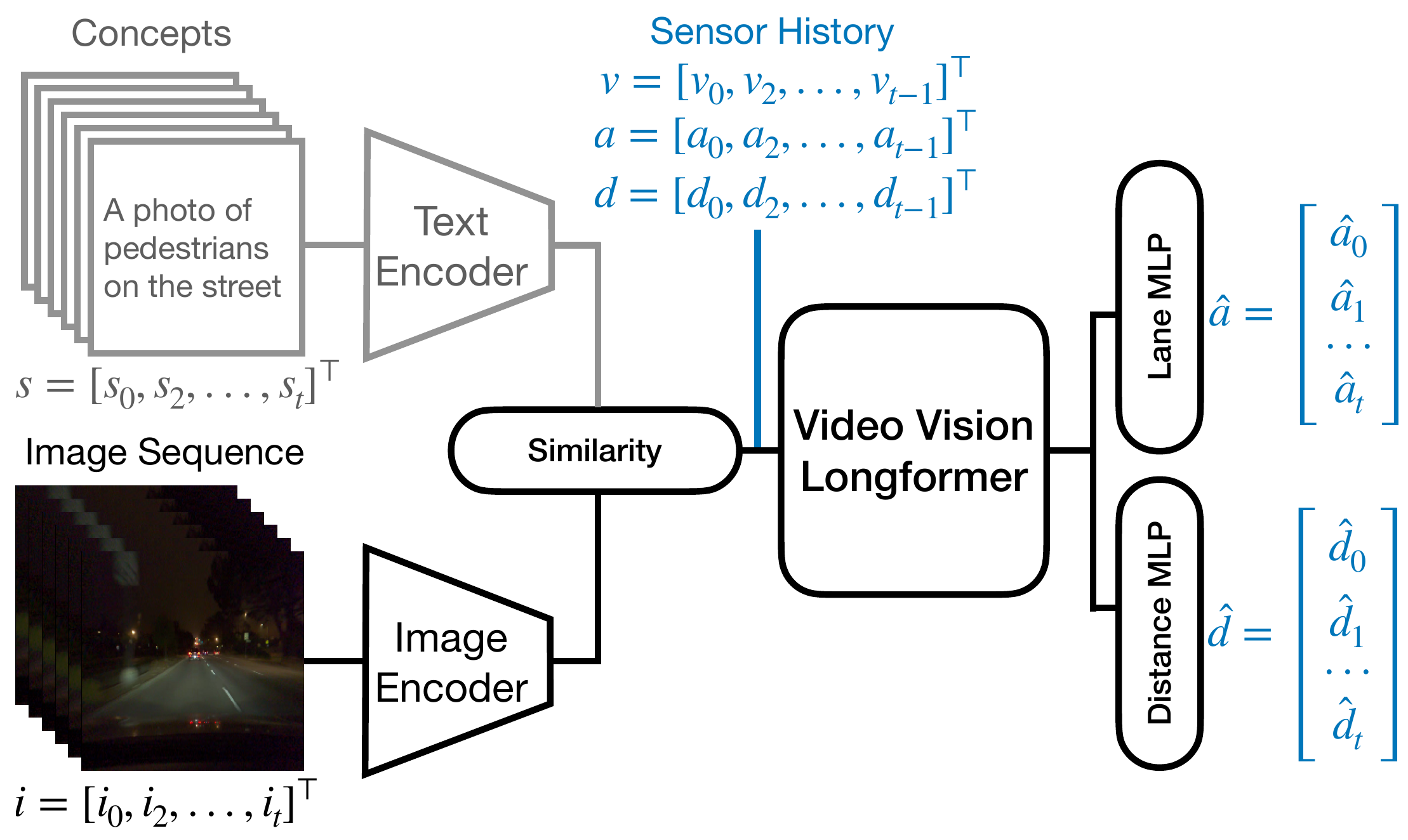}
        \caption{Pipeline of the interpretable concept bottleneck control command prediction. $v,a$ and $d$ denote the sensor history of speed, steering angle and leading vehicle distance.}
        \label{fig:pipeline}
\end{figure}
\section{Methods}
Consider predicting a target value $y \in \mathbb{R}$ from input $x \in \mathbb{R}^d$, while trying to gain reasoning $c$ for the prediction of the target value. 
That is, we observe training points $\{(x^{(i)}, y^{(i)})\}^n_{i=1}$, and we want to determine $y^{(i)}, c^{(i)}$ where $c^{(i)} \in \mathbb{R}^k$ is a vector of $k$ concepts. We consider bottleneck models of the form $f(g(x))$, where $g : \mathbb{R}^d \rightarrow \mathbb{R}^k$ maps an input $x$ into the concept space (``clear skies", ``a car in the lane ahead in close proximity", etc.), and $f : \mathbb{R}^k \rightarrow \mathbb{R}$ maps concepts into a final prediction (\eg forward distance is 40 meters). These types of models are called concept bottleneck models \cite{koh2020concept, oikarinen2023label} because the control command prediction $\hat{y} = f(g(x))$ relies on the input $x$ through the bottleneck prediction $\hat{c} = g(x)$.
\subsection{Image Feature Backbone}
Our method takes inspiration from video vision transformer networks \cite{neimark2021video, vivit}. Typically, spatial backbones take on the function of $g : \mathbb{R}^d \rightarrow \mathbb{R}^k$ that maps an input $x$ into the latent (un-interpretable) feature space, (\eg Neimark et al. \cite{neimark2021video} use the video vision transformer from Arnab et al. \cite{vivit} as a latent feature bottleneck). However, we incorporate explainability through concept bottleneck models \cite{koh2020concept}. This 
%\textcolor{blue}{
pre-trained
%} 
conceptual spatial backbone operates as a learned feature extraction module to determine sequential decisions or control commands. 
%\textcolor{blue}{Additional experiments with joint conceptual spatial backbone training compared to our separate training process is left for future work due to the lack of concept-labeled data in the automated driving domain,} 
We compare this model with traditional convolutional- or transformer-based methods  \cite{https://doi.org/10.48550/arxiv.1512.03385,ging2020coot, yang2022recurring, vivit, DBLP:journals/corr/HeZRS15}. Let these features $F_{\text{input}}$ denote the input feature to the subsequent sequential evaluation component (Longformer). 

\subsection{Concept Bottleneck}\label{cb}
When replacing the feature backbone with a concept bottleneck model we construct driving scenarios $s$. These scenarios are supposed to describe scenes and encode contextual information about the driving scenario in natural language. A scenario captures factors such as road conditions, traffic density, and weather conditions. To obtain those scenarios, we leverage two concept curation methods. First, we use the generative capabilities of GPT-3.5 \cite{ouyang2022training} to create diverse driving scenarios. We specifically ask the language model to provide scenarios as described in the following, starting with very general scenarios and subsequently generating more fine-grained scene explanations.
\begin{itemize}
    \item List scenarios that could occur in traffic  starting each sentence with \{\emph{a photo of ...}\}
    \item List scenarios that could occur in traffic with respect to \{\emph{weather; traffic participants, lane changing, highway driving, city driving}\} starting each sentence with \{\emph{a photo of ...}\}
\end{itemize}
Just like Radford et al. \cite{radford2021learning}, we follow the template of  \{\emph{a photo of ...}\} (\eg \emph{a photo of a car driving on a highway}) as a default, as it has been shown that the performance of specific concept bottleneck models can be increased this way \cite{radford2021learning}. 
 
These generated scenarios are then combined with a subset of existing human-created scene descriptions from the NuScenes dataset \cite{caesar2020nuscenes}. We transformed them into the same pattern from Radford et al. \cite{radford2021learning}. This allows us to enrich the dataset with other diverse driving contexts, \eg ``pedestrians" or ``workers on the street", as well as compare different concept curation methods. We then manually filter the set for duplicates. The specific construction of the concept space $\mathbb{S}$ is domain-specific and can be customized for other driving-related domains. To the best of our knowledge, this is the first captured driving-related concept bottleneck, and we release these scenarios and code upon publication. 

In the concept bottleneck model $g$, we encode the image features $x$ using an image encoder $g_{\text{image}} : \mathbb{R}^d \rightarrow \mathbb{R}^l$ and scenarios $s \in \mathbb{S}$ using a text encoder $g_{\text{text}} : \mathbb{R}^s \rightarrow \mathbb{R}^l$ \cite{radford2021learning}. For each image, we can then measure the similarity between the embedding $g_\text{image}(x)$ and the scenarios $g_\text{text}(s)$ employing cosine similarity:
\begin{equation}
    \text{sim}_\text{cos}(x,s) = \frac{{g_\text{image}(x) \cdot g_\text{text}(s)}}{{\|g_\text{image}(x)\| \|g_\text{text}(s)\|}}
\end{equation}
where $\cdot$ represents the dot product, and $|\cdot|$ denotes the Euclidean norm to get an indication of what is happening in image frames from the driving sequence.
\subsection{Temporal Encoder}
Video vision transformers encode visual features in a temporal manner with a transformer architecture that was originally developed for natural language processing \cite{vaswani2017attention}. This acts as our regression module $f : \mathbb{R}^l \rightarrow \mathbb{R}$ for each frame encoded with $g$. The attention mechanism in a transformer neural network is given by
\begin{equation}
    \text{Attention}(Q, K, V) = \text{softmax}\left(\frac{QK^T}{\sqrt{d}}\right)V
\end{equation}
where where $Q$, $K$, and $V$ are the query, key, and value matrices, respectively, and $d$ is the dimensionality of the key vectors. The softmax function normalizes the attention weights.
Due to the original transformer attention complexity of $O(n^2)$ , a Longformer architecture with a sliding window attention \cite{beltagy2020longformer} is useful to reduce computational overhead \cite{neimark2021video}. Given a sliding window size $w$ and sequence length $n$, its complexity is reduced to  $O(w \times n)$ \cite{beltagy2020longformer}. To attend to other time steps in the sequence, we use a window attention size of eight frames. The sequence of feature vectors from the backbone and the sensor history of previous vehicle speed $v$, steering angles $a$ and distance to leading vehicles $d$ for each captured frame is fed to the Longformer model as shown in \cref{fig:pipeline}. We prepend a special token ([CLS]) at the beginning of the feature sequence. The Longformer maintains global attention on that special [CLS] token. After propagating the sequence through the Longformer layers, we use the final state of the features related to this classification token as the final representation of the video and apply it to the given regression task head to learn the control commands through a linear regression head. Each output from the temporal encoder is processed with an MLP head to provide a final predicted value. The MLP head contains two linear layers with a GELU \cite{DBLP:journals/corr/HendrycksG16} non-linearity and dropout \cite{srivastava2014dropout} between them. The input token representation is processed with Layer normalization. We use one MLP for each task trained separately, and two MLPs for the multi-task setup. The idea for evaluating multi-task setups is that a person might be more careful in their overall driving behavior for both tasks, so that the two tasks could benefit from being trained together. We train our models with Root Mean Squared Error (RMSE) loss $\mathcal{L} = \sqrt{\frac{\sum_{i=1}^N(f(g(x))-y)^2}{N}}$ with $g(x) = \textit{sim}(x,s)$.
\section{Experiments}
\subsection{Data}
For comparative evaluation, we employ two datasets consisting of diverse driving scenarios captured from real-world driving situations. The datasets encompass a wide range of environmental conditions, traffic scenarios, and driver behaviors to ensure generalizability of our findings.

\textbf{Comma2k19.}
We explore the Comma 2k19 dataset \cite{DBLP:journals/corr/abs-1812-05752}, which captures commute scenarios with different features, \eg visual images, CAN data (\eg steering wheel angle), and radar data (distance to preceding vehicle) in the San Francisco Bay Area. The Comma data mostly consists of highway scenarios, and their captured sequences are comparatively long compared to other datasets. In total, Comma 2k19 has 100GB of data from 33 hours of driving. In this work, we use a 25GB subset of the data. The data was captured at 20fps and was subsampled to 4fps to reduce redundancy for training. All data sequences are one minute long, but continuous driving sequences per session ranged between 3 and 13 minutes. For our purposes, each driving sequence consists of 240 samples.
%\begin{table}[b]
%    \centering
%    \begin{tabular}{c|ccc}
%    \toprule
%        & \multicolumn{3}{c}{\textbf{Explainability}}\\
%        Method & Visual & Linguistic & Temporal\\
%        \midrule
%        ResNet+Longformer & \cmark & \xmark& \cmark\\
%        ViT+Longformer & \cmark & \xmark & \cmark\\
%        CLIP+Longformer &\cmark & \xmark & \cmark\\
%        Concept+Longformer &\cmark & \cmark& \cmark\\
%        \bottomrule
%    \end{tabular}
%    \caption{Regular methods can achieve visual explanation, \eg through gradient visualization like \cite{selvaraju2016grad}. The Longformer approach provides possibility for temporal explanations through its attention mechanism. Concept bottlenecks can provide additional linguistic explainability.}
%    \label{tab:my_label}
%\end{table}

\begin{table*}[t]
    \centering
\scalebox{0.78}{
\begin{tabular}{lll|lll}
\toprule
 Dataset & Model &  Feat. Size & a-MAE & d-MAE & (a,d)-MAE  \\
\midrule
%Comma &   Swin \cite{jin2023adapt} &   X      & X     &    X      &  -\\
Comma &   ResNet+GapFormer\cite{sachdeva22gapformer}\footnote{The original work only uses sensor input data.} &         512 &      0.08 &         0.28 &  - \\
   Comma &     CLIP+Longformer &         512 &      0.03 &         7.95 &  [0.22, 8.97] \\
   Comma &      ViT+Longformer&         768 &      0.06 &         5.23 &   [0.8, 6.08] \\
   Comma &   ResNet+Longformer &         512 &      0.03 &         3.79 &  [0.37, 4.11] \\
   Comma &     Concept (Full)+Longformer&         643 &       0.7 &         0.97 &  [0.36, 1.83] \\
   Comma &      ResNet+Concept (Full)+Longformer & 1,155      &    0.37  &    2.43     &  [2.15, 1.74] \\

   \midrule
%NuScenes &   Swin \cite{jin2023adapt} &   X       & X     &    X      &  -\\
NuScenes &   ResNet+GapFormer\cite{sachdeva22gapformer} &         512 &      0.57 &         0.74 &  - \\
NuScenes &     CLIP+Longformer&         512 &      0.57 &         5.46 &  [3.51, 3.47] \\
NuScenes &      ViT+Longformer&         768 &      3.75 &         1.31 & [0.44, 16.62] \\
NuScenes &   ResNet+Longformer &         512 &      5.87 &         26.5 & [9.47, 43.81] \\
NuScenes &     Concept (Full)+Longformer&         643 &      1.89 &           4.21 &           [0.36, 6.65] \\
NuScenes &   ResNet+Concept (Full)+Longformer&   1,155       & 0.97     &    4.8      &  [2.46, 4.26]\\

\bottomrule
\end{tabular}}
    \caption{Mean Absolute Error (MAE) performance of different models on the downstream task of steering angle (a) and distance (d) prediction in a single and multi-task setting, compared to the inherently explainable concept bottleneck model.}
    \label{tab:res_multitask}
\end{table*}

\begin{figure*}
     \centering
     \begin{subfigure}[b]{0.23\textwidth}
         \centering
         \includegraphics[width=\textwidth]{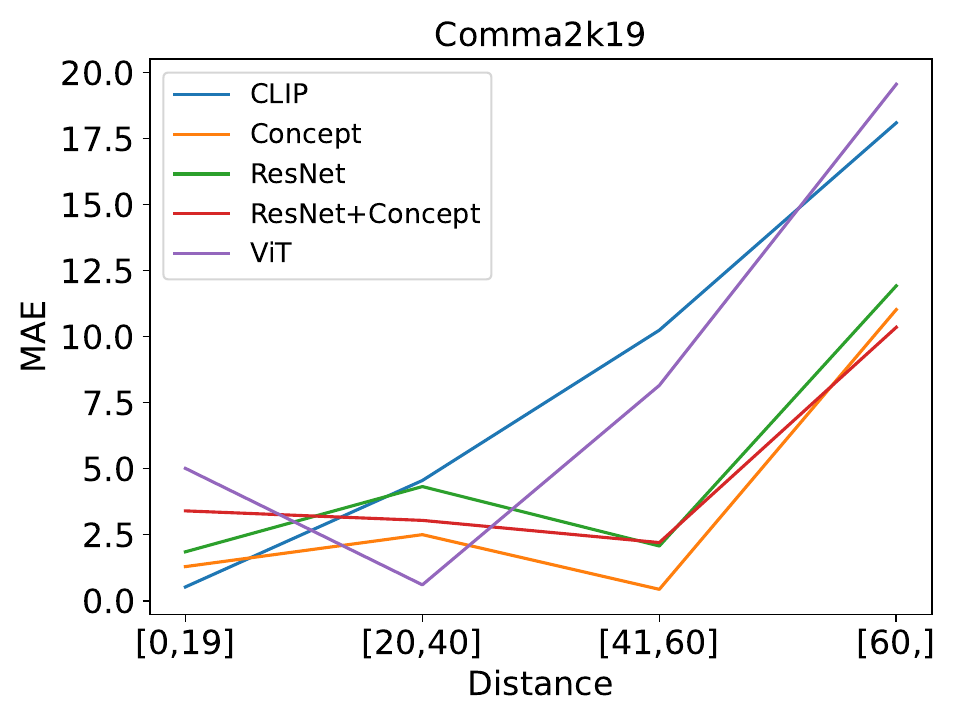}
     \end{subfigure}
     \begin{subfigure}[b]{0.23\textwidth}
         \centering
         \includegraphics[width=\textwidth]{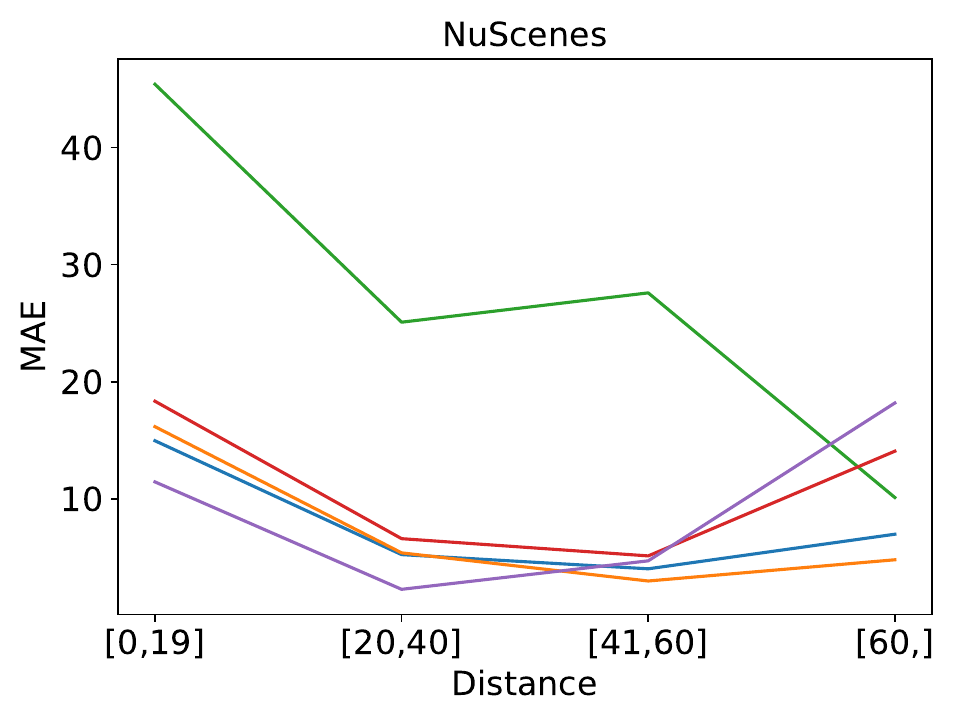}
     \end{subfigure}
     \begin{subfigure}[b]{0.23\textwidth}
         \centering
         \includegraphics[width=\textwidth]{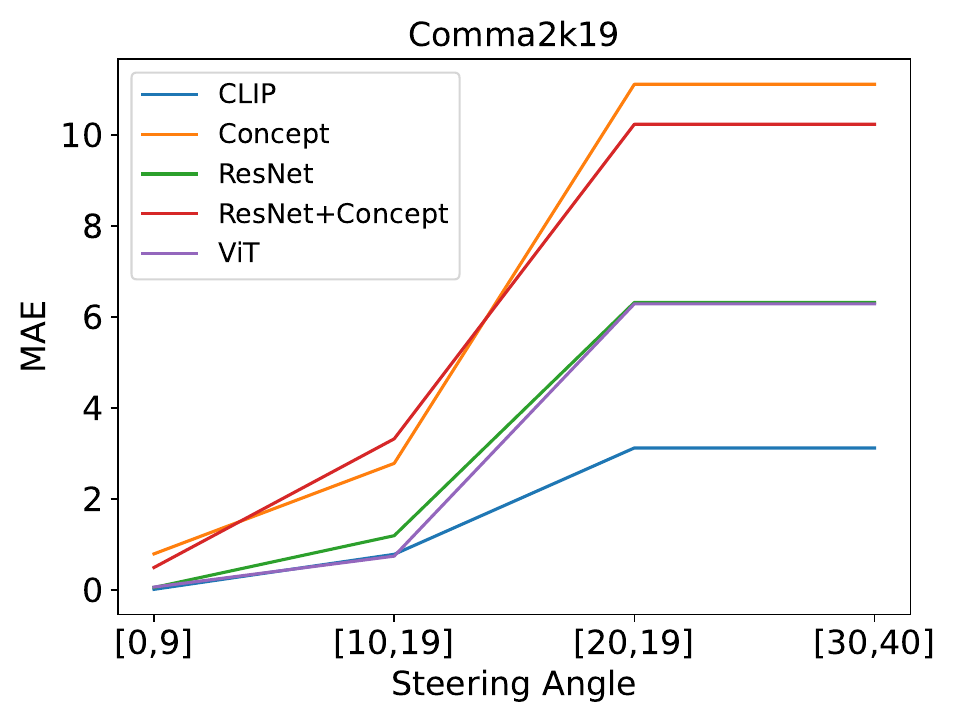}
     \end{subfigure}
     \begin{subfigure}[b]{0.23\textwidth}
         \centering
         \includegraphics[width=\textwidth]{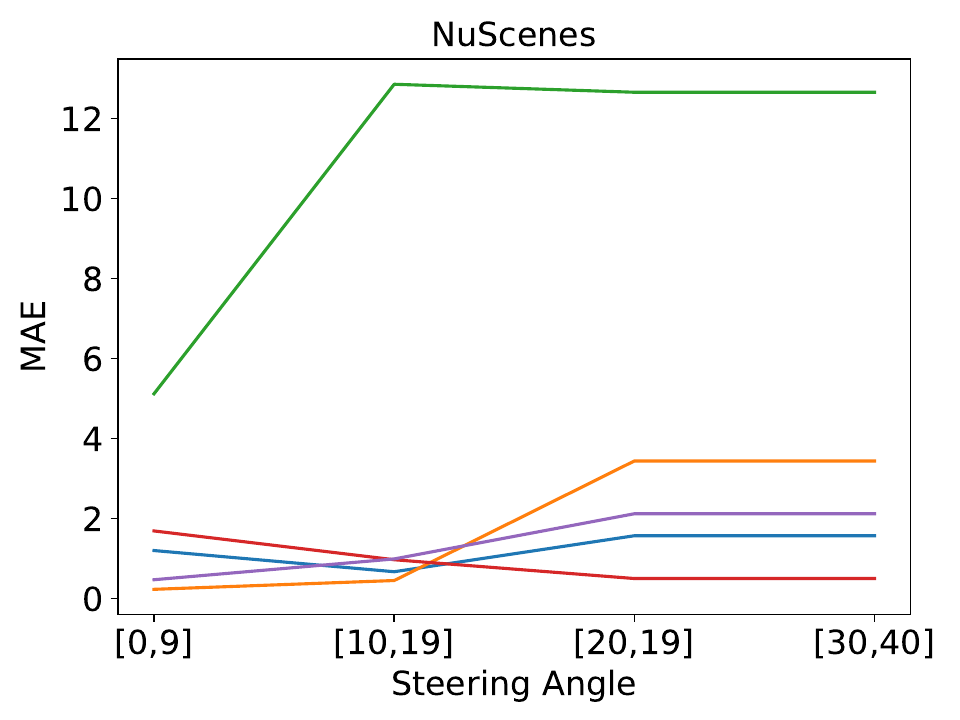}
     \end{subfigure}
     \caption{Error analysis of different model backbones and tasks. We see fewer absolute error on smaller ground truth forward distances and steering angles. This intuitively makes sense as the visual information is clearer for small gaps (e.g. when directly following a leading vehicle), compared to longer distances. Similarly, learning small steering angles is easier, \eg for lane keeping in highway driving, compared to turning on an intersection. Our concept bottleneck model continually performs similarly or better to other approaches.}
        \label{fig:error}
\end{figure*}
\textbf{NuScenes.}
The NuScenes dataset \cite{caesar2020nuscenes} is collected using a fleet of autonomous vehicles equipped with lidar, radar, cameras, and ego-motion sensors, and is designed for the development and evaluation of perception, planning, and control algorithms. With data captured in various urban driving scenarios across multiple cities, the NuScenes dataset provides researchers and developers with a range of environments and traffic conditions to analyze. The dataset includes annotations for each sensor modality, including 3D bounding boxes, as well as natural language scene descriptions, enabling algorithm development and evaluation in a structured manner. Each scene of the dataset consists of 20 seconds and is resampled at 1fps. The descriptions serve as ground truth for our the concept bottlenecks. We use a subset of 250 scenes for evaluation of our method. 

The two datasets serve different purposes. (1) the comma dataset provides long driving sequences to learn potential interventions on highway scenarios, that can be connected to explanatory driving behavior. For example, we might like to explain a change in leading-vehicle gap, occurring due to a change of scenario (\eg someone cut in the front lane), versus changed user preferences. (2) the NuScenes dataset with its natural language scene annotations can evaluate the explanatory abilities of our model. These two datasets capture a wide variety of scenarios in city and highway driving. For both datasets we resize all frames to $224 \times 224$ pixels. We exclude distances over 70m for our evaluation, as we empirically evaluated that distances beyond this threshold contain little visual information useful for gap prediction (\eg no leading vehicle present). We use a 0.85/0.05/0.1 train/val/test split.
\begin{figure*}
     \centering
     \begin{subfigure}[]{0.48\textwidth}
         \centering
         \includegraphics[width=\textwidth]{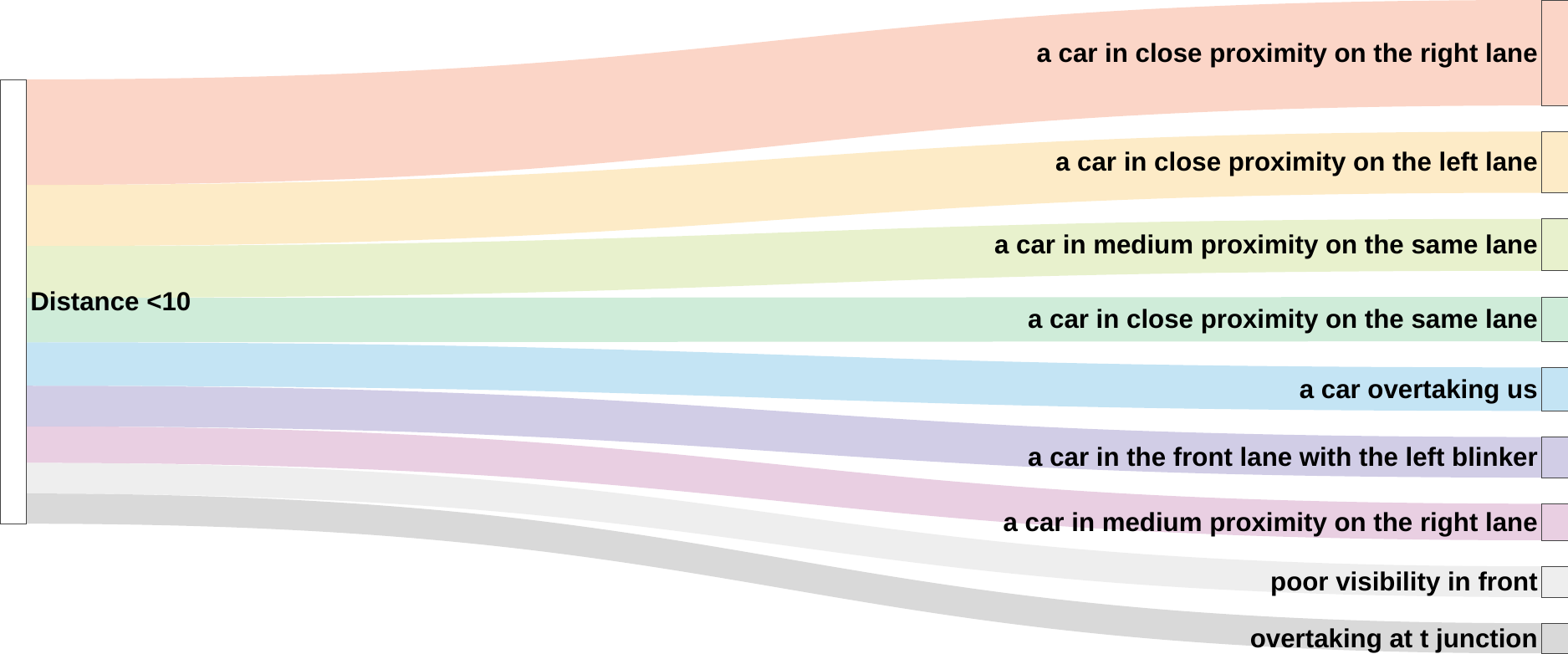}
     \end{subfigure}
     \hfill
     \begin{subfigure}[]{0.48\textwidth}
         \centering
         \includegraphics[width=\textwidth]{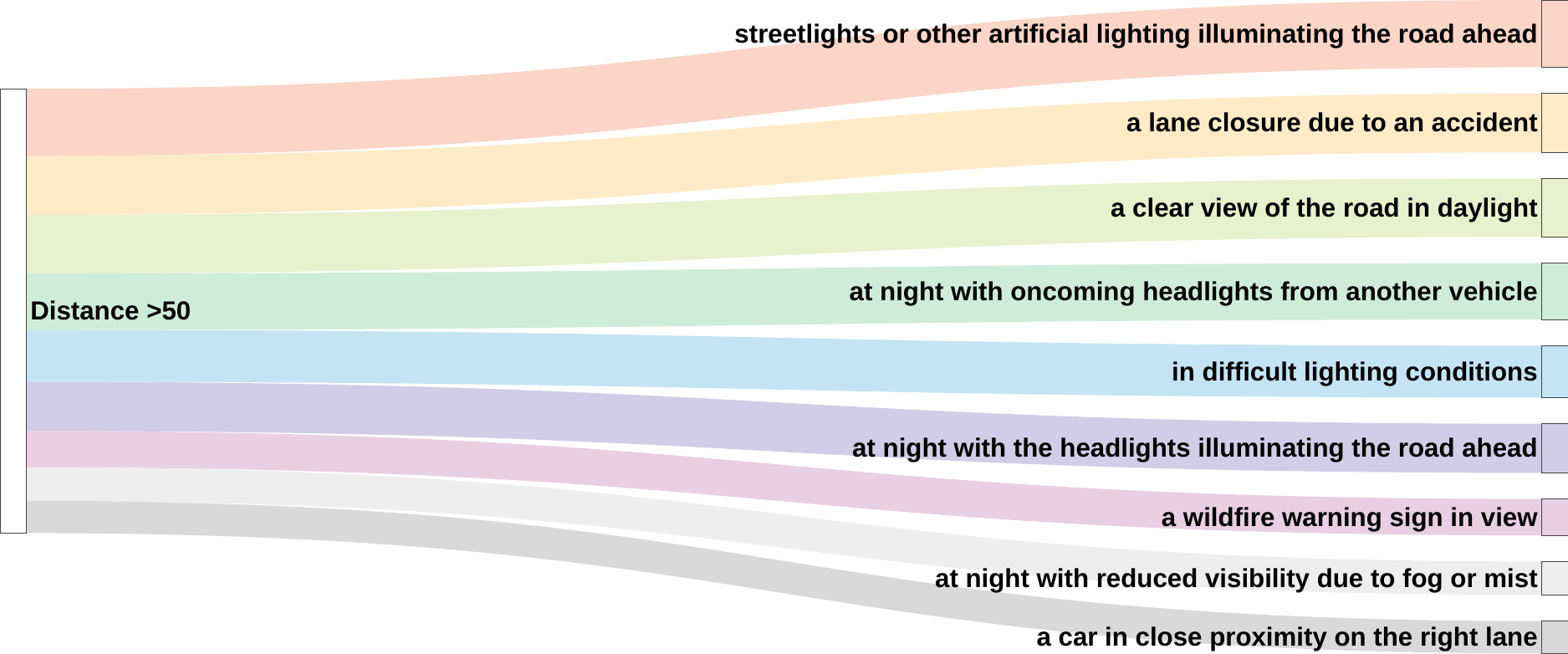}
     \end{subfigure}
     \caption{Visualization of explanation capabilities of our model to determine reasons drivers keep short forward distances (\eg distance $<10$ meter or longer distances (\eg distance $>50$ meter). Height of the lines indicates fraction of top-10 predictions. }
        \label{fig:distance}
\end{figure*} 
\subsection{Backbones}
To identify how explainable concept bottleneck models perform compared to standard methods, we conduct an analysis of different backbone models. We evaluate the performance of ResNet-18 \cite{https://doi.org/10.48550/arxiv.1512.03385}, Vision Transformer \cite{vivit}, and CLIP \cite{radford2021learning} backbones for our single- and multi-task control command prediction task. ResNet-18 \cite{https://doi.org/10.48550/arxiv.1512.03385}, with its deep architecture and skip connections, has been a benchmark backbone model in computer vision. Vision Transformer (ViT) \cite{vivit} replace convolutional layers with self-attention mechanisms, with an ability to capture global dependencies. We investigate the performance of the CLIP image backbone, which is another transfomer-based backbone, but typically its ViT-based image encoder is combined with a transformer-based text encoder. We analyze its effectiveness in capturing visual-semantic representations with only its image-based encoder. Concept bottlenecks can provide additional linguistic explainability without requiring additional language generation models (such as LSTMs in \cite{kim2018textual}).
\section{Results}
\subsection{Control-Command Prediction Performance}\label{sec:ccperf}
We evaluate the performance of concept bottleneck models as interpretable feature extractors for downstream tasks. \cref{tab:res_multitask} presents the Mean Absolute Error (MAE) performance of different black-box backbones compared to the concept bottleneck model on the tasks of steering angle and distance prediction in both single and multi-task settings. It can be observed that the concept bottleneck models achieve a competitive MAE across different datasets. In particular, for the Comma dataset, the concept bottleneck model with a feature size of 643 obtains with a MAE of 0.7 for angle prediction and 0.97 for distance prediction. Similarly, for the NuScenes dataset, the concept bottleneck model with a feature size of 643 achieves a MAE of 1.89 for angle prediction and 4.21 for distance prediction. These results indicate that concept bottleneck models exhibit good performance as interpretable feature extractors for downstream tasks. We see no significant difference in performance of our concept bottleneck approach between single- and multi-task setups. In \cref{fig:error}, we show error based on different ground-truth magnitudes. We observe that concept bottleneck models as feature extractors can lead to better performance of control command prediction, while convolution-based approaches may fail to learn the task (on the NuScenes dataset). However, when the visual properties are connected more strongly to the task (e.g. for gap prediction, compared to steering angle prediction), we see an increased utility and performance. 
%\textcolor{blue}{
We also find that our prediction procedure has an average model inference latency over 100 runs of 0.1 seconds (excluding data processing), and system throughput (including data processing) of 1 per second using an NVIDIA RTX A6000 GPU for long sequences (240 frames on Comma2k19) and 2 per second for short sequences (20 frames on NuScenes).
%}

\subsection{Scene Explanation Capabilities}
\begin{table}[]
    \centering
    \scalebox{0.9}{
    \begin{tabular}{c|c|c}
    \toprule
        Concept Curation & Comma2k19 & NuScenes\\
        \midrule
         Human &  1.77 & 3.93\\
         GPT-3.5 & 0.89 & 2.02\\
    \bottomrule
    \end{tabular}}
    \caption{Comparison of concepts that were created by humans (adapted from \cite{caesar2020nuscenes}) versus curated from GPT-3.5 \cite{ouyang2022training} for predicting lead vehicle distance. We can see that automatically curated concepts can perform better in terms of distance MAE compared to human curated concepts.}
    \label{tab:curation}
\end{table}
By employing the concept bottleneck model, we can analyze and interpret the factors contributing to larger or smaller gaps to leading vehicles. The interpretability of the concept bottleneck model allows us to understand the underlying causes behind these gap variations, shedding light on human decision-making processes in relation to preceding vehicles. In \cref{fig:distance}, we see that smaller gaps are typically associated with a prediction of \emph{``vehicles in  close and medium proximity"} or \emph{``cars in the front lane"}. On the other hand, large distances are associated with the prediction of \emph{``a clear view"}, \emph{``difficult lighting conditions"}, or \emph{``at night"}. Intuitively a driver might keep a larger leading distance at night, and shorter distances in \eg a traffic jam.
\begin{figure*}
     \centering
     \begin{subfigure}[b]{0.84\textwidth}
         \centering
         \includegraphics[width=\textwidth]{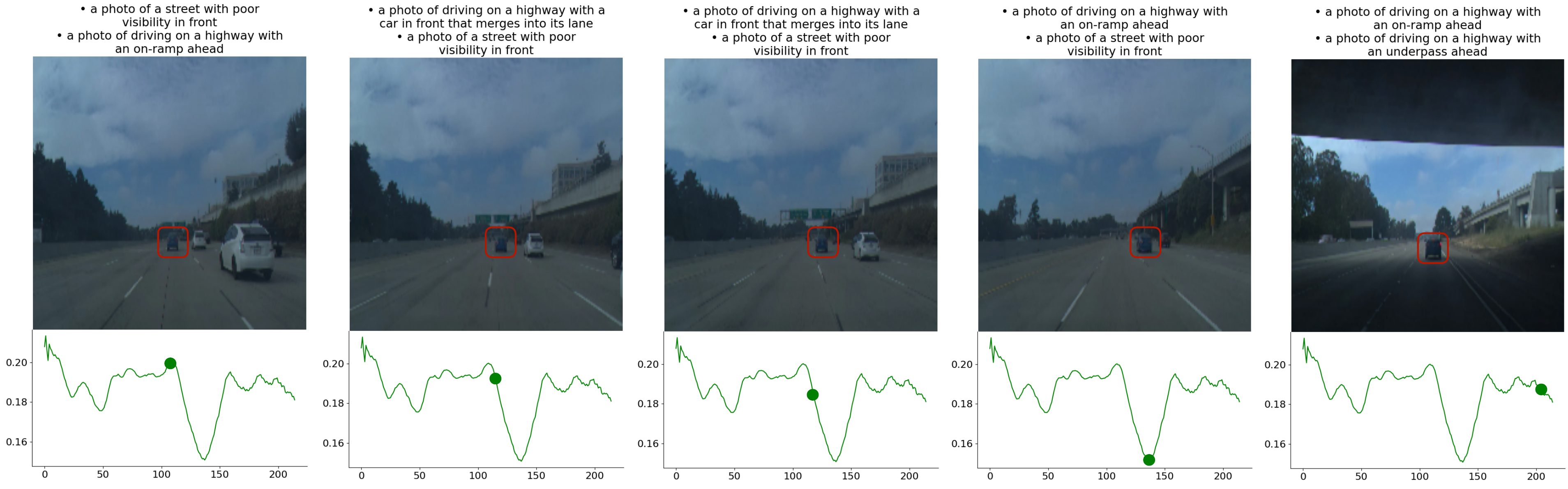}
     \end{subfigure}
     \begin{subfigure}[b]{0.84\textwidth}
         \centering
         \includegraphics[width=\textwidth]{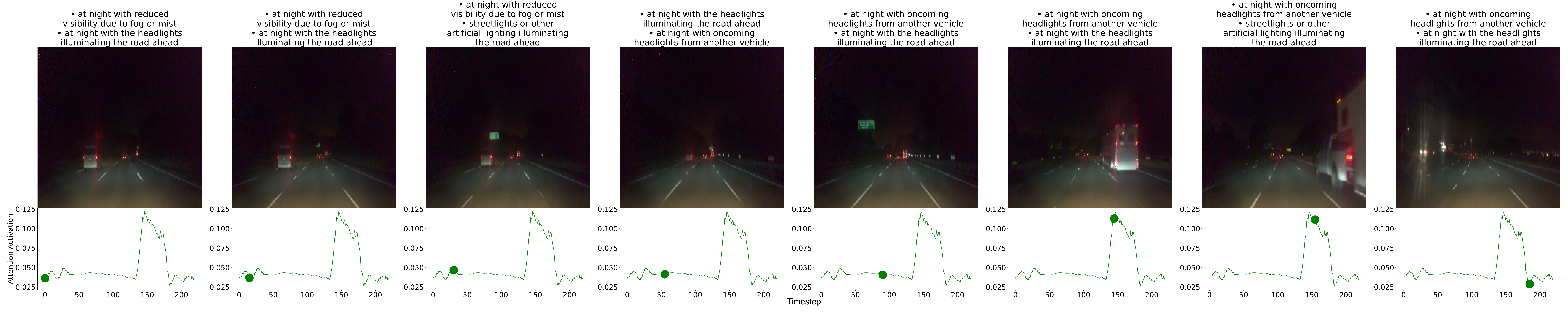}
     \end{subfigure}
     \begin{subfigure}[b]{0.84\textwidth}
         \centering
         \includegraphics[width=\textwidth]{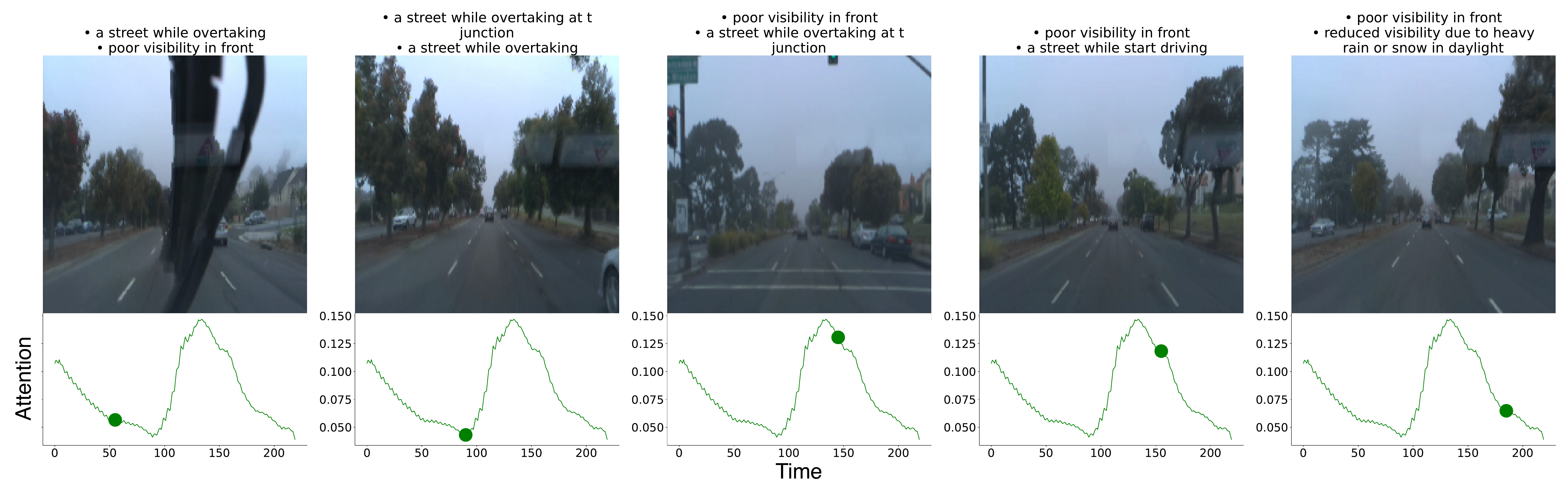}
     \end{subfigure}
     \caption{Three scenarios from the Comma dataset for gap prediction with scenario explanations from the concept bottleneck and their attention values (y-axis) over time (x-axis) at particular points in time (green dot). We observe that the Longformer attention is a good indicator for when interventions might happen. For example, we see a leading car changing the lane, leading to attention drop (top) or a scenario of passing at a t-junction, leading to a spike (bottom); the ego vehicle passes a trailer vehicle on the right, leading to an attention spike (middle). }
        \label{fig:atten}
\end{figure*}

%\textcolor{blue}{
To quantitatively evaluate the effectiveness of explainability through the concept bottleneck, we design a human evaluation study of 50 images per dataset. We evaluate the top-10 concepts for each frame and extract the top-3 occurring concepts over 20 frames. We present each short video with the predicted concepts to three human crowdworkers and ask them how many of the concepts are correct. When we aggregate the worker votes by majority vote, we find that 94\% of the top-3 concept predictions have at least one correct concept for NuScenes and 90\% for Comma2k19. A fine-grained evaluation of individual reviews (not majority voted) shows that (for NuScenes/Comma) 9\%/15\% of instances are labeled as having no correct concepts, 30\%/32\% as having one correct concept, 38\%/34\% have two correct concepts and 23\%/19\% have all top-3 concepts correct.
%}

Additionally we calculate the common content words between NuScenes scene descriptions and concept predictions, such that the NuScenes descriptions serve as a form of ground-truth. We consider the top-3 concept predictions for each frame and then the top-3 concept predictions for the entire scene, and remove any stopwords to only evaluate relevant content words for each scene and scenario from the concept bottleneck. By considering the top-3 predicted concepts, we are able to correctly explain 81\% of the scenes. When considering the top-1 concept prediction, we can explain 76\% of all scenes accurately. This demonstrates the capability of our approach to effectively explain the content of scenes by leveraging concept predictions and their intersection with scene descriptions. In \cref{fig:atten}, we also show predictions of the concept model on driving scenarios.
\subsection{Concept Curation}
Concept curation plays a vital role in building a comprehensive understanding of the automated driving domain. Traditionally, it has relied on human experts who bring their expertise and domain knowledge to the curation process. Humans can provide nuanced insights, contextual understanding, and connections between different concepts based on their experience, but they are also costly and subjective. Human curation can also be time-consuming, limited by individual biases, and susceptible to errors or omissions. We evaluate a randomly selected subset of 270 human created concepts, adapted with the template from \cref{cb} 
%\textcolor{blue}{
from the scene descriptions of the NuScenes dataset. These textual descriptions were made by expert annotators to add captions for each scene (e.g.: ``Wait at intersection, peds on sidewalk, bicycle crossing, jaywalker'') \cite{caesar2020nuscenes}. We additionally evaluate 270 generated concepts by GPT 3.5 similar to \cite{ouyang2022training}, yielding sentences like ``driving on a highway with an overpass overhead''.
%} 
Our results show that human curation is not better compared to concept curation obtained by large language models \cref{tab:curation}; on the Comma2k19 dataset we achieve a distance MAE of 0.89 for GPT curated concepts, and 1.77 for human curated concepts on the Comma dataset, and MAE of 2.02 for GPT curated concepts, and 3.93 for human curated concepts on the NuScenes data %\textcolor{blue}{ 
\cref{tab:curation}.
%}
\subsection{Does attention matter?}
We provide an analysis of three scenarios extracted from the Comma2k19 dataset, focusing on the gap prediction task (\cref{fig:atten}). These scenarios are accompanied by explanations generated from the concept bottleneck model, offering insights into the underlying factors influencing the observed gap variations. In our analysis, we investigate the role of the Longformer attention mechanism as a valuable indicator for identifying instances where interventions might occur. By examining the attention, we can discern patterns and changes in the scene that may prompt user intervention. In the first scenario, we observe the ego car changing lanes. This maneuver often requires careful monitoring and potential intervention from the driver. By examining the attention distribution captured by the Longformer, we notice a significant drop in attention at the moment when the ego car changes the lane and has a free lane ahead, and an attention increase when it is back in a lane behind a vehicle. This attention drop suggests that the concept bottleneck model correctly identifies this critical event and recognizes the reduced relevance of certain features in the scene. In the second scenario, we encounter a situation where the ego vehicle passes a trailer vehicle on the right side. This scenario often demands extra caution and anticipation from the driver, as the presence of large vehicles can impact the driving environment. Analyzing the attention distribution, we observe a spike in attention when the trailer vehicle enters the scene. This attention spike indicates that the model successfully captures the significance of this change in the scenery, identifying the trailer vehicle as a prominent object that requires increased attention.  The third scenario involves a noteworthy change in the driving environment on a T-junction while driving in the rain. As we analyze the attention patterns, a spike in attention occurs when our vehicle encounters the junction situation and the attention serves as an indicator for the change in the scenery. The concept bottleneck model, through its attention mechanism, effectively recognizes and highlights these critical moments, and can explain these scenarios through its bottleneck activations.

%\textcolor{blue}{
We evaluate the Longformer attention to observe whether it can be used to select when to reveal a concept to a user. If a particular part of a sequence in (semi-) autonomous driving is of relevance, indicated by the attention, it can be used to decide if an intervention from the autonomous car is required, if the user should take over and provide reasond why (given the concept explanation).
%}
%Our work finds an interesting relationship between attention and visual concepts, but a more exhaustive evaluation of this phenomenon should be tackled in future work.}
\begin{table}[]
    \centering
    \scalebox{0.9}{
    \begin{tabular}{l|l|lll}
    \toprule
         Dataset & Feat. Size & a-MAE & d-MAE \\
         \midrule
         Comma & 24 & 0.38 & 3.48 \\
         Comma & 48 & 0.3 & 1.15 \\
         Comma & 100 & \textbf{0.27} & \textbf{0.22} \\
         Comma & 300 & 0.43 & 0.6 \\
         Comma & Full & 0.7 & 0.97 \\
         \midrule
         NuScenes & 24 & 1.49 & 1.85 \\
         NuScenes & 48 & 0.60 & \textbf{0.02} \\
         NuScenes & 100 & \textbf{0.52} & 0.51\\
         NuScenes & 300 & 2.27 & 4.01\\
         NuScenes & Full & 1.89 & 4.21  \\
         \toprule
    \end{tabular}}
    \caption{Bottleneck size (randomly selected from full (643) bottleneck) versus command control prediction performance. We see that bottleneck size seems to have a significant impact on performance, with a sweet spot at 100 concepts.}
    \label{tab:ablation}
\end{table}
\subsection{Does Bottleneck Size Matter?}
We investigate the impact of bottleneck size on the performance, to evaluate how the size of the bottleneck affects the accuracy of control command predictions. The ablation study varies the size of the bottleneck while keeping all other factors constant.
The ablation study results are summarized in \cref{tab:ablation}. The feature size denotes the number of concepts in the bottleneck layer, which were randomly drawn from all possible 643 scenarios. We observe that as the bottleneck size increases, both steering angle MAE and distance MAE decrease. For Comma data with a bottleneck size of 24, the steering angle MAE is 0.38 and distance MAE is 3.48. With a bottleneck size of 100, the steering angle MAE decreases to 0.27, and the distance MAE drops to 0.22. Interestingly, further increasing the bottleneck size to 300 resulted in worse performance. We observe a similar tendency for the NuScenes dataset: increasing the bottleneck size leads to improved prediction accuracy. With a bottleneck size of 24, the steering angle MAE is 1.49 and distance MAE is 1.85. Increasing the bottleneck size to 100 reduces both steering angle MAE (0.52) and distance MAE (0.51) with performance degradation for larger concepts. The findings indicate a impact of bottleneck size on the prediction accuracy, with a ``sweet spot" at a bottleneck size of 100 concepts. There are different reasons for the performance benefits with smaller concept sizes. %
Previous work shows that it is possible to achieve good performance with smaller concept spaces \cite{yan2023learning} and that correlated concept spaces can be an issue \cite{heidemann2023concept}. We conjecture that the performance benefit in our work for a smaller concept space may be based on (1) multiple concepts in the original concept set having only small deviations, which means that we might achieve the same or better results when excluding them. For example, the difference between the concept ``a pedestrian crossing''; ``a pedestrian crossing crosswalk''; ``a pedestrian crossing traffic light'' can be subtle. (2) the image size of $224 \times 224$ pixels does not allow for fine-grained concept granularity. For example different street signs like ``a curve sign''; ``a steep hill sign''; ``a winding road sign'' can be too fine-grained to be visible. 
%}%Test time interventions of concepts (like in \cite{koh2020concept}) would be interesting to perform for further insights into concept interactions for different .}
%We would like to note that a specific human concept curation against those nuances and correlations might be fruitful in future work. In addition, test time interventions of concepts (like in \cite{koh2020concept}) would be interesting to perform in future work for insights into concept interactions.}

\section{Conclusion}
This study validates the effectiveness of concept bottleneck models for explainability in sequential settings for automated driving. Our work leverages a concept bottleneck model and Longformer sequential processing unit within a control command prediction setup and we show competitive performance to standard black-box approaches. Using our method, we identify and explain factors contributing to changes in driving behavior both visually through linguistic explanation as well as temporally through transformer attention. This can explain \eg changes in forward distance to a leading vehicle, and enable a deeper understanding of the decision-making processes in automated driving. Our model demonstrates effectiveness in explaining scene content, which can serve as a baseline for future work aligning linguistic, visual and temporal explanations. For example, future work could explore more use-cases (such as speed prediction), fuse more modalities into the prediction procedure, or analyse bottleneck uncertainty (e.g. with test-time interventions) in more detail.

%%%%%%%%% REFERENCES
{\small
\bibliographystyle{ieee_fullname}
\bibliography{egbib}
}

\end{document}